\documentclass[conference]{IEEEtran}
\IEEEoverridecommandlockouts

\usepackage{cite}
\usepackage{amsmath,amssymb,amsfonts}
\usepackage{algorithmic}
\usepackage{graphicx}
\usepackage{textcomp}
\usepackage{xcolor}
\usepackage[export]{adjustbox}
\usepackage{subcaption}
\usepackage{makecell}
\usepackage{url}
\usepackage{booktabs}

\def\BibTeX{{\rm B\kern-.05em{\sc i\kern-.025em b}\kern-.08em
    T\kern-.1667em\lower.7ex\hbox{E}\kern-.125emX}}

\begin{document}

\title{Design and Integration of Thermal and Vibrotactile Feedback for Lifelike Touch in Social Robots}

\author{Jacqueline Borgstedt\IEEEauthorrefmark{1}\IEEEauthorrefmark{2}, Jake Bhattacharyya\IEEEauthorrefmark{2}, Matteo Iovino\IEEEauthorrefmark{1}, Frank E. Pollick,\IEEEauthorrefmark{2} and Stephen Brewster\IEEEauthorrefmark{2}

\thanks{\IEEEauthorrefmark{1}ETH Zürich, Switzerland}
\thanks{\IEEEauthorrefmark{2}University of Glasgow, Scotland}}

\maketitle

\begin{abstract}
Zoomorphic Socially Assistive Robots (SARs) offer an alternative source of social touch for individuals who cannot access animal companionship. However, current SARs provide only limited, passive touch-based interactions and lack the rich haptic cues, such as warmth, heartbeat or purring, that are characteristic of human-animal touch. This limits their ability to evoke emotionally engaging, life-like physical interactions.

We present a multimodal tactile prototype, which was used to augment the established PARO robot,  integrating thermal and vibrotactile feedback to simulate feeling biophysiological signals. A flexible heating interface delivers body-like warmth, while embedded actuators generate heartbeat-like rhythms and continuous purring sensations. These cues were iteratively designed and calibrated with input from users and haptics experts. We outline the design process and offer reproducible guidelines to support the development of emotionally resonant and biologically plausible touch interactions with SARs.
\end{abstract}

\section{Introduction}

Chronic exposure to stress is widely recognized to negatively affect physical and psychological well-being, leading to the onset and deterioration of disorders related to anxiety, depression, elevated blood pressure, or hypertension~\cite{OConnor2021,Shields2017}. Stress mitigation can be achieved by interacting with friends and family, especially through touch, which is reported to reduce human stress levels~\cite{Coan2006,Jakubiak2019} and improve stress resilience~\cite{Dagnino-Subiabre2022}. 
The benefits of social touch are not only limited to humans, but extend to other species such as canines, with studies reporting the effectiveness of human-animal interaction in calming participants during stress-provoking situations~\cite{Nimer2007, Eckstein2020_touch_stress}. Although human-animal interaction is often regarded as a viable substitute for human-human touch, its accessibility is limited for many demographics due to,  for example, allergies, housing restrictions, or financial constraints~\cite{Haggerty2017}.

\begin{figure}[t]
  \centering
  \includegraphics[width=\linewidth]{figs/MultiModalPrototype.png}
  \caption{\textit{Multimodal Prototype}, integrating an electric heating pad controlled via a manual temperature regulator. It includes multiple vibrotactile actuators that simulate both purring and heartbeat sensations.}
  \label{fig:multimodal_thermal_prototype}
\end{figure}

To overcome this limitation, zoomorphic robots have been proposed as an alternative source of social touch and animal-like companionship, and as a potential technological intervention to support well-being~\cite{Eckstein2020_touch_stress, Chen2019, packheiser2024_touch_interventions}. These types of robots are known as Socially Assistive Robots (SARs) and are designed primarily to engage with users socially by offering emotional support, companionship, or motivation through interaction~\cite{seifer2005_social_robotics_definition}. \textit{Zoomorphic} SARs, such as PARO~\cite{Geva2020, Wada2005, Aminuddin2017} or the Huggable~\cite{Jeong2015}, are designed to engage users in physical, social interactions by leveraging an animal-like embodiment. By engaging users in familiar touch-based interactions such as petting, stroking, or hugging, these robots aim to simulate the comforting aspects of human-animal relationships.

Despite the increase in commercial availability of zoomorphic SARs (e.g., PARO, Quooboo, Joy for All pets, Aibo), their functionalities remain limited and often rely on passive touch-based interactions, where the robot only reacts after being touched in designated areas and remains idle otherwise. Generally, touch-based interactions with robots are limited to feeling the robots’ shell or movements, while touching an animal involves complex haptic cues such as feeling its body temperature, heartbeat, or purring. The lack of diverse haptic interactions with SARs limit the experience of users who seek believable, biologically plausible, life-like touch interactions that emulate the feeling of interacting with a real animal.

To address these limitations and building on prior findings in~\cite{borgstedt2022hot, borgstedt2024soothing},  a multimodal prototype of the PARO robot was developed that integrates thermal and vibrotactile feedback (as shown in Figure~\ref{fig:multimodal_thermal_prototype}). The thermal interface is designed so that it is flexible and can be wrapped around PARO's torso, providing a broader thermal contact area than the one proposed in~\cite{borgstedt2022hot}, resulting in a more realistic or biologically plausible  display of body temperature. Vibrotactile cues are delivered by actuators directly embedded in PARO's body and have been iteratively designed to generate a purring and heartbeat cue that achieves a more realistic and emotionally resonant feedback than ~\cite{borgstedt2024soothing}.  The design process outlined in this paper are not limited to the PARO robot alone but is, in principle, applicable to other zoomorphic robot embodiments.

The main contributions of this work are:
\begin{itemize}
    \item An iterative, user- and expert-informed design of thermal and vibrotactile cues that enhance comfort, perceived realism, and emotional engagement with a zoomorphic robot;
    \item A detailed, generalizable design methodology for integrating multimodal haptic cues, enabling reproducibility and adoption by other researchers for diverse SAR platforms;
    \item A novel multimodal prototype that augments an established, evidence-based robot in an accessible and cost-effective manner, demonstrating the potential to extend existing interventions with richer, life-like interactions.
\end{itemize}

\section{Background and Related Work}

Having motivated the needs for and benefits of zoomorphic SARs in the introduction, this section is dedicated to reviewing existing platforms, describing their functionalities and limitations. Zoomorphic robots are designed to facilitate human-robot interaction by evoking human-animal touch dynamics, primarily involving physical interactions such as petting, hugging, or playing. SARs such as PARO register users' touch through touch-sensitive sensors, facilitating multi-point touch-based interactions by mimicking animal-like behavior such as movements. 

Haptic technologies are systems that communicate information through the sense of touch, typically complementing or sometimes replacing visual and auditory modalities. These interfaces use various forms of mechanical stimulation, including pressure, vibration, force, and temperature, to enable users to receive information or to interact with an interface~\cite{hayward2004_haptics}. Vibrotactile feedback in Human-Computer Interaction research has been used to communicate emotions to users ~\cite{rantala2013touchgestures,Ju2021emotioncommunicationvibro, Wilson2017}, and  has been shown to enhance body awareness, emotion regulation~\cite{dobrushina2024interoceptive}, or to influence internal physiological states~\cite{zhou2020_heartbeat_haptics}. Critically, research on physiological entrainment suggests that internal bodily rhythms (e.g., blood pressure, breathing, heart rate) can passively synchronize with external rhythms~\cite{bernardi2009entrainment}. These findings could be leveraged in the context of social robotics in care, such that displays of external vibrotactile rhythms (e.g. a robot emitting a vibrotactile heartbeat) could be used as a stress intervention, an approach that has been shown to be successful in the space of wearable technologies~\cite{Azevedo2017} and explored and shown promise in early works on robotics by~\cite{borgstedt2024soothing}.

Among human-human interactions, perceiving another person's body temperature is a fundamental component of human touch experiences~\cite{hertenstein2002touch_heat}, and temperature has been shown to play a critical role in how affectionate touch is perceived~\cite{ackerley2014_touch_warmth}. A similar pattern has been observed in interactions with robots, where users interacting with warm robots during stressful situations reported that the robots felt more trustworthy, friendly, and human-like~\cite{Nie2012}. Thermal cues have also been shown to improve perceived social connection, emotional engagement~\cite{Park2014}, and closeness~\cite{Nakanishi_warm_robot_hand}. While these studies demonstrate the promise of thermal and vibrotactile feedback to enhance social presence, most existing platforms implement these cues in isolation, offer limited temperature options, or are designed without iterative user input. In contrast, our work integrates thermal and vibrotactile feedback multimodally, with customizable temperature settings and cues embedded directly within the robot. Crucially, the design of these cues was iteratively refined based on both user feedback and haptic expert input, creating a more realistic, emotionally resonant, and user-informed interaction. By documenting this process, we provide a generalizable methodology for designing multimodal haptic cues that can be applied to other zoomorphic robots beyond PARO, supporting both reproducibility and wider adoption in HRI research.

PARO, a robotic baby seal, is the most extensively studied and commercially used platform in elderly care. It features multi-modal sensing (e.g., tactile, sound, temperature, posture) and responds to physical and verbal interactions with life-like movements and vocalizations. Research consistently links its use to reduced stress and loneliness, and improved mood and quality of life in older adults~\cite{Hung2019_paro_review}. However, its lack of thermal and haptic feedback has been noted as a limitation in achieving more immersive interactions~\cite{borgstedt2024soothing}. To address this, simple prototypes have been proposed: for example,~\cite{borgstedt2022hot} added thermal elements hidden in PARO’s ears, which could be calibrated to users’ preferred temperatures. While positively received, participants noted that the limited surface area and constrained placement restricted natural interaction and reduced the immersive quality of the thermal cues. Similarly, a unimodal vibrotactile prototype featuring heartbeat cues~\cite{borgstedt2024soothing} showed promise in enhancing users’ emotional connection, but the cues were generated by an external device rather than being embedded within the robot, limiting perceived realism.

Life-like haptic cues have also been explored in entirely new research prototypes such as Teo~\cite{Bonarini2016}, Probo~\cite{Saldien2008}, Snugglebot~\cite{Bates2020}, the Haptic Creature~\cite{Yohanan2011design}, and Huggiebot 2.0~\cite{Block2021}, which integrate warm, soft materials, vibrotactile cues, breath-like motion, and multi-modal sensing to enrich physical interactions. While these designs are innovative, their being entirely new devices makes it difficult to determine whether lifelike haptic cues truly enhance the therapeutic effects of companion robots, and replication by other groups is often challenging. In contrast, our work augments PARO with embedded thermal and vibrotactile cues, allowing direct comparisons between baseline PARO and our multimodal prototype. This approach not only provides rigorous insight into the added value of lifelike haptic cues but also ensures that our design is accessible and replicable by other researchers who use PARO. Moreover, our prototype was iteratively refined with input from users and haptic experts, ensuring that the cues reflect user preferences rather than assumptions such as “users will like warm temperatures.” By building on a well-established, evidence-based intervention, our work provides a reproducible, user-centered methodology for integrating multimodal haptic feedback into zoomorphic robots, enabling both scientific evaluation and practical extension of existing SARs.

\begin{figure}[t]
  \centering
  \includegraphics[width=\linewidth]{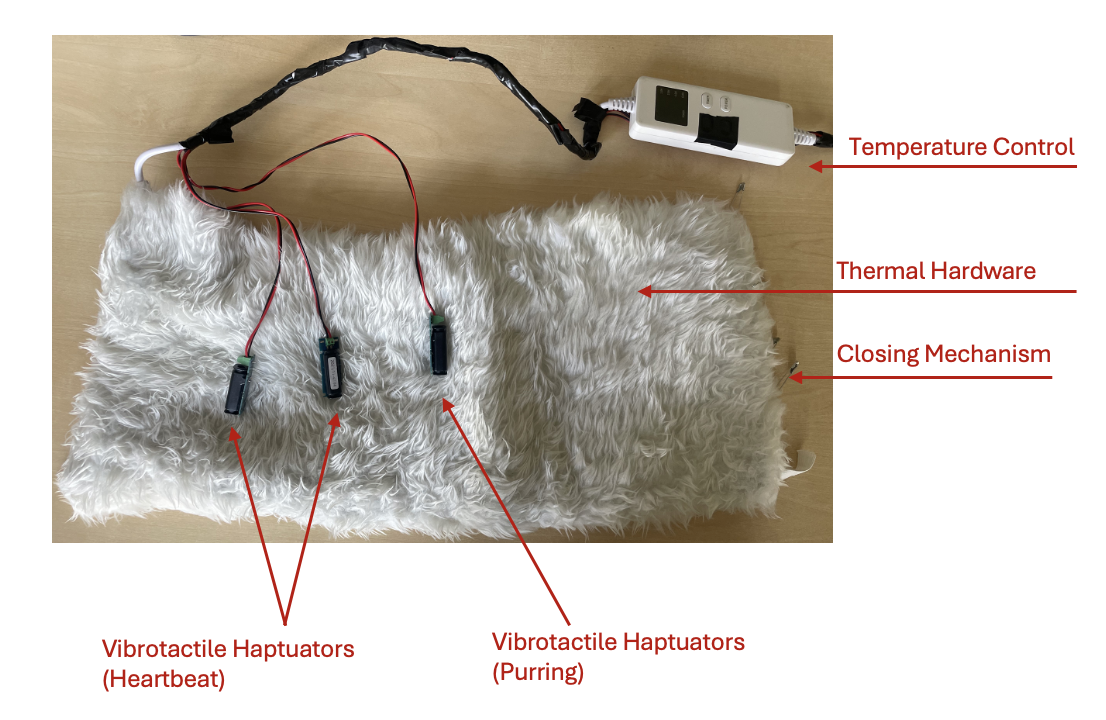}
  \caption{\textit{Thermal and Vibrotactile Interfaces}. The electric heating pad is sewn into a layer of soft white faux fur and controlled via a manual temperature regulator. The vibrotactile actuators are used to simulate both purring and heartbeat sensations.}
  \label{fig:thermal_interface}
\end{figure}

\section{Thermal Interface}

The goal of the novel thermal hardware was to deliver perceptible warmth across a larger surface area in a reliable and safe manner, building on a simple uni-modal prototype proposed by~\cite{borgstedt2022hot}, who used a Peltier heat sink to deliver both hot and cold thermal cues. A range of commercially available thermal devices, including portable heated scarves, socks, gloves, and insoles, were tested and evaluated based on the following criteria: flexibility to wrap around PARO's torso, surface coverage, heating power, safety, and the potential for discreet integration to camouflage the added hardware. While active cooling was initially considered as a potential feature, it was ultimately excluded due to both technical and practical constraints.

This decision was further supported by preliminary data collected by \cite{borgstedt2022hot}, who reported that although the full range of users temperature preferences spanned from $4^\circ C$ to $36^\circ C$, the majority of participants preferred temperatures in the neutral to warm range\--particularly $32^\circ C$ and $34^\circ C$. Neutral or unheated configurations remain viable for users who do not prefer warmth, while future research may explore integrated, seamless cooling solutions more appropriate for scalable or commercial deployment. 

The thermal cues of the proposed interface are delivered by the OneAmg Electric Heat Pad\footnote{\url{https://www.amazon.co.uk/OneAmg-Settings-Washable-Electric-Abdominal/dp/B0B8SK9RXN}}, which features a grid-distributed heating wire that allows the pad to rapidly heat up. Being a commercial product, the original heating pad is integrated in a gray fleece covering with built-in Velcro straps for ease-of-use. In our interface (shown in Figure~\ref{fig:thermal_interface}), the straps were removed and the gray fleece cover was sewn into a white machine-washable synthetic fur that closely mimicked PARO’s original texture and visual appearance. This modification maintained the illusion that warmth was emanating from the robot itself, rather than from an attached external device. The pad was then secured around the torso of the robot using non-permanent fasteners that allowed for easy removal and replacement between sessions. 

This adapted off-the-shelf interface offers several technical advantages:
\begin{itemize}
 \item \textbf{Adjustable Temperature Settings}: The heating pad features six pre-set temperature levels, with measured surface temperatures ranging from approximately $33^\circ C$ to $43^\circ C$, using a FLIR infrared camera. Although not distributed uniformly across the whole surface, temperatures were more moderate and stable in the central region, which is where the users typically interacted with the robot. By consequence, we will refer to measurements taken from this region in the remainder of the paper  (shown in Table~\ref{tab:thermal_settings}).
 \item \textbf{Flexible and Safe to Bend}: Unlike many electric heating pads, the OneAmg model is designed to be soft and flexible, making it suitable for wrapping around curved surfaces like PARO’s torso without compromising functionality or safety.
 \item \textbf{Washable Fabric with Waterproof Connector}: The heating pad features a waterproof power connector and a removable controller, enabling machine washing of the pad itself when the cable is detached. This feature is particularly valuable in research and healthcare settings where hygiene and repeated use are concerns.
 \item \textbf{Built-in Timer Settings}: The device includes a timer with programmable shut-off durations, ensuring user and robot safety during prolonged sessions and preventing overheating.
\end{itemize}

With this solution, we achieve a balance between functional thermal feedback, user safety, and realistic interaction, while also promoting replicability, and cost-effectiveness.

\subsection{Evaluation}

\begin{table}[t]
\centering
\caption{Summary of the settings considered in the evaluation of the thermal interface.}
\label{tab:thermal_settings}
\begin{tabular}{ccc}
\toprule
\makecell{\textbf{Setting}} & \makecell{\textbf{Setting Specification} \\ \textbf{(Temperature Range in $^\circ C$)}} & \makecell{\textbf{High Level} \\ \textbf{Description}} \\
\midrule
1 & $28^\circ C$ & Baseline \\
2 & $35-36^\circ C$ & Mild Warmth \\
3 & $38-39^\circ C$ & Warm \\
4 & $40-43^\circ C$ & Very Warm \\
\bottomrule
\end{tabular}
\end{table}

To evaluate the thermal hardware system described in the previous section we gathered data on users’ perceived comfort and the biological plausibility of the novel device. To this end, we conducted a pilot study with 20 participants (10 men, 10 women) aged 24–42 years (M = 28.65, Median = 27), all of whom were junior researchers affiliated with our university. All sessions took place in a quiet office space under stable ambient temperature conditions (Min = $21.4^\circ C$, Median = $22.44^\circ C$, Max = $23.5^\circ C$), reflecting realistic indoor environmental variation.

The goal of this study was to explore which temperature settings were perceived as comfortable during interaction with the PARO robot. Although the commercially available heating pad included six preset temperature levels, informal pretesting revealed that the differences between adjacent settings were too subtle for users to reliably distinguish. Because our aim was to understand how comfort and perceived realism varies in response to clearly different temperature levels, we selected only three temperature levels that we tested alongside a neutral baseline condition, representing PARO's natural surface temperature at room temperature (reported in Table~\ref{tab:thermal_settings}). 

The settings were presented in increasing order of temperature. We chose this order to overcome the slow cooling characteristics of the thermal hardware, which made randomization impractical. If at any point a participant experienced and reported discomfort, the setting would not be increased. However, no participant reported feeling uncomfortable. Each temperature setting was turned on for a total of five minutes. During this time the robot was placed on the participant’s lap. The participant was instructed to interact with PARO as they wished and they were free to pet, hold, or rest their hands on the robot. Participants could also engage in light conversation with the researcher, send messages on their laptop, or perform other casual activities while sitting with the robot on their lap. The informal nature of this data collection session was intentional to  approximate real-world interaction contexts of using the robot while studying. 

After experiencing a given temperature setting for 5 minutes, each participant completed a short questionnaire including the following items: ``I felt comfortable with this temperature for the entire duration'' and ``PARO’s body temperature feels like a real animal''. Participants rated each statement on a 5-point Likert scale (1 = Strongly Disagree, 5 = Strongly Agree). Furthermore, participants were asked to briefly describe their experience, thoughts, and concerns regarding the temperature setting. After experiencing all temperature conditions, participants completed a final questionnaire, which asked them about their perceived comfort levels, if they experienced any discomfort, and then asked to rank the temperature settings in terms of perceived comfort (1 = most preferred, 4 = least preferred). Finally participants were asked to describe and explain their subjective experiences across the temperature settings.

\begin{figure*}[t]
    \centering
    \begin{subfigure}[b]{0.45\linewidth}
        \centering
        \includegraphics[width=\linewidth]{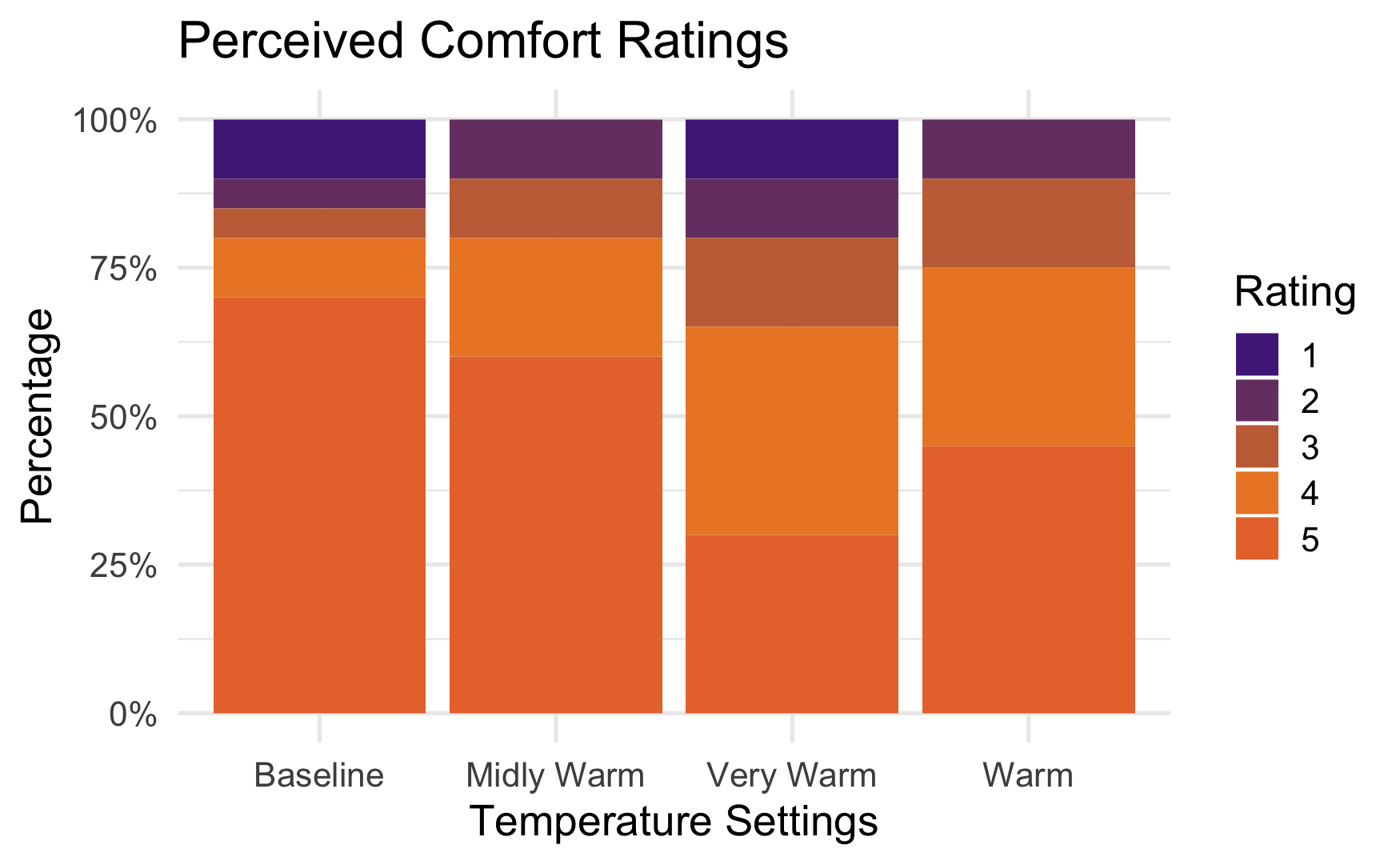}
        \label{fig:perceived_comfort}
    \end{subfigure}
    \hfill
    \begin{subfigure}[b]{0.45\linewidth}
        \centering
        \includegraphics[width=\linewidth]{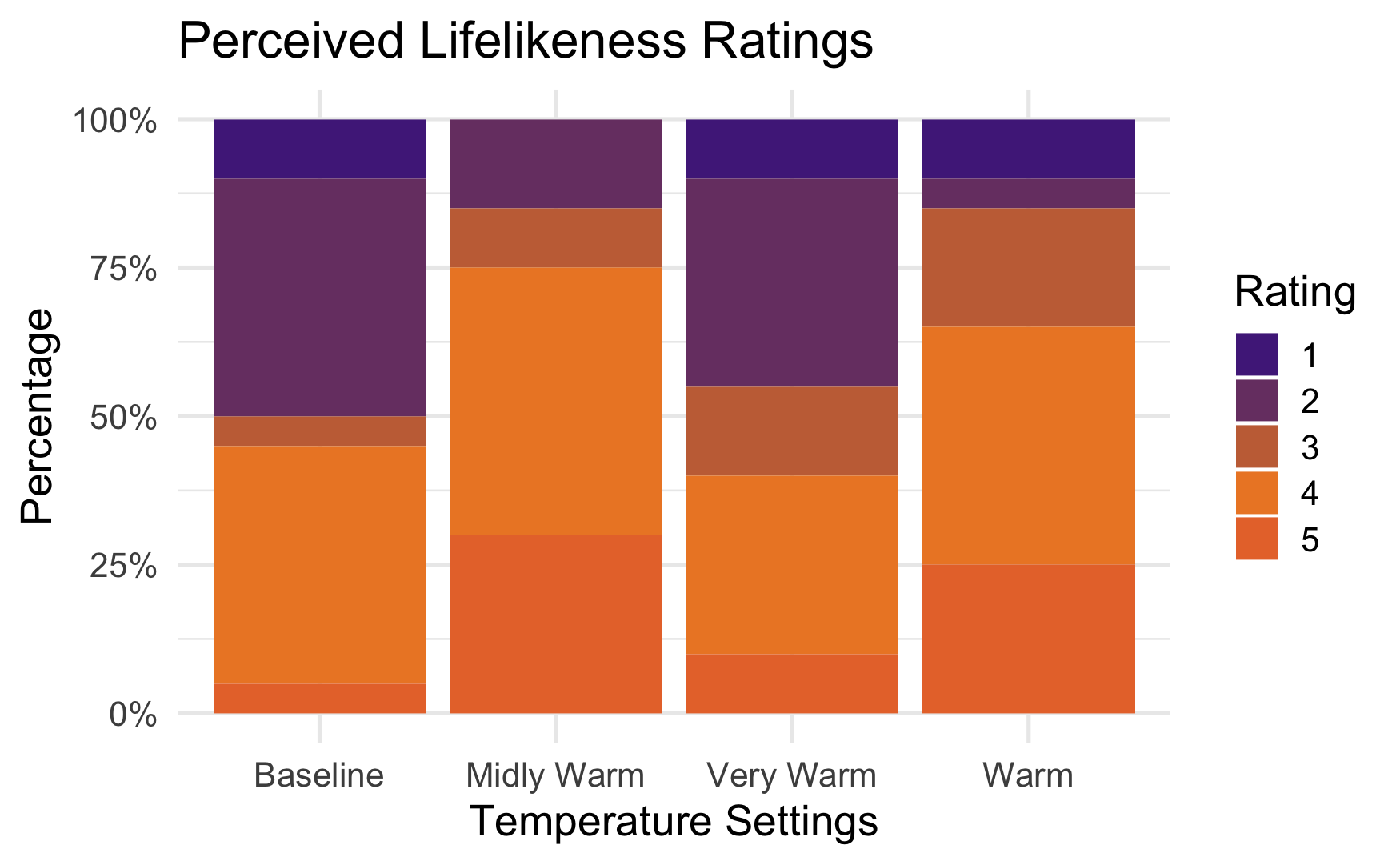}
        \label{fig:perceived_realism}
    \end{subfigure}
    \caption{Stacked bar charts of perceived comfort and realism ratings (1 = lowest, 5 = highest) across four temperature settings: Baseline ($28^\circ C$), Mildly Warm ($35-36^\circ C$) , Warm ($38-39^\circ C$), and Very Warm ($40-43^\circ C$). Each bar represents 100\% of responses, segmented by comfort/realism rating.}
    \label{fig:comfort_realism_temperature}
\end{figure*}

\paragraph{\textbf{Analysis on Perceived Comfort}}

Participants generally perceived all temperature conditions as sufficiently comfortable, as shown in Figure~\ref{fig:comfort_realism_temperature} . Mean comfort ratings were similar across conditions: Setting~2 ($35$-$36^\circ$C) had the highest mean (M = 4.30, Median = 5), followed closely by Setting~1 (neutral, M = 4.25, Median = 5) and Setting~4 ($38$-$39^\circ$C; M = 4.10, Median = 4). Setting~6 ($40$-$43^\circ$C) received the lowest ratings (M = 3.65, Median = 4), with a few participants reporting mild discomfort. However, a linear mixed-effects model with temperature setting as a fixed effect and participant ID as a random effect, combined with an Aligned Rank Transform (ART)~\cite{Wobbrock2011} to account for the ordinal nature of the data, revealed no significant effect of temperature condition on comfort ($F(3) = 2.42$, $p = .075$). This suggests that, although minor numerical differences were observed descriptively, the inferential results indicate that participants experienced comparable levels of comfort across all four temperature conditions.

\paragraph{\textbf{Analysis on Realism}}

To evaluate whether the perceived realism of the robot's body temperature varied across different thermal conditions, we fit a linear mixed effects model, which included temperature setting as a fixed effect and participant ID as a random effect:
\begin{equation*}
    \mbox{realism} \sim \mbox{temp\_setting} + (1\vert \mbox{ID}).
\end{equation*}
The same analysis strategy as in the previous section was employed. This revealed a statistically significant main effect of temperature on perceived realism, $F(3) = 3.84$, $p = .014$, with a partial eta-squared ($\eta^2$) of .17 ($95\% CI [0.02, 1.00]$), indicating a large effect size.

\textit{Post hoc} comparisons, using the ART-C method~\cite{Elkin2021} showed that Setting~2, corresponding to the mild warmth condition ($35-36^\circ C$), was associated with significantly higher perceived realism compared to Setting~1, the neutral condition ($28^\circ C$) with no active heating. No other pairwise differences between settings reached statistical significance. Mean realism ratings further support this pattern: participants rated Setting~2 as the most realistic ($35-36^\circ C$, M = 3.90, Median = 4.0), followed by Setting~4 ($38-39^\circ C$; M = 3.65, Median = 4.0), while both the Baseline Setting~1 (M = 2.90, Median = 2.5) and the hottest Setting~6 ($40-43^\circ C$; M = 2.95, Median = 3.0) received noticeably lower ratings as shown in Figure~\ref{fig:comfort_realism_temperature} right.

\paragraph{\textbf{Qualitative Analysis of Temperature Perception}}

After experiencing each temperature setting, participants provided text-based qualitative feedback regarding their experience of each temperature setting. This qualitative data was analysed using~\cite{Braun2006}'s 6-stage Framework for Thematic Analysis. However, in the context of iterative design and since preferences for temperature are likely to vary across individuals, we additionally quantified the number of participants endorsing a given perception (e.g., realism, discomfort, or emotional connection), allowing the evaluation of which thermal settings were experienced as most life-like or acceptable by the majority.

In the interest of space, we do not report quotes from the participants but instead we provide the themes within the responses grouped by temperature. However a more detailed analysis of results can be found in the supplementary materials\footnote{\url{https://github.com/JacquelineBorgstedt/multimodal_prototype}}.
\begin{itemize}
 \item \textbf{\textit{Baseline} }(Setting 1) : the findings suggest that while the absence of temperature does not induce discomfort, it fails to support life-likeness or emotional connection. The most prominent theme, reported by 9 out 20 participants, was a desire for warmth, often motivated by a wish for increased realism or emotional engagement. Warmth was not universally required for comfort, but its absence was consistently interpreted as a lack of biological or social presence.
 \item \textbf{\textit{Mild Warmth}} (Setting 2) : this setting was most often described as striking the ideal balance between physical comfort and enhanced realism. 8 participants explicitly highlighted enhanced realism, often in direct contrast to the baseline condition. It was warm enough to evoke biological realism and encourage closeness, yet mild enough to remain comfortable for 18 out of 20 users. It functioned not only as a physical cue but as a socially meaningful signal, prompting participants to relate to the robot as more than a mechanical object.
 \item \textbf{\textit{Warm}} (Setting 3):  this setting represented a perceptual threshold: a total of 11 participants stated that the temperature was comfortable but borderline too hot. While still broadly acceptable for most participants, the increase in warmth challenged the boundaries of realism and comfort for some, revealing how small shifts in temperature can affect the perception of life-likeness and comfort.
 \item \textbf{\textit{Very Warm}} (Setting 4): this setting surpassed a threshold where warmth ceased to support affective realism and instead disrupted it. While 2 participants framed the experience positively, interpreting the warmth as soothing,  11 participants experienced it as excessive. The findings underscore the importance of calibrating upper limits in thermal design, ensuring warmth enhances engagement without compromising believability or comfort.
\end{itemize}

These findings reinforce and nuance the statistical results: Setting~2 (mild warmth) offered the best balance between comfort, realism, and emotional resonance. As temperatures increased, comfort declined and users perceived the robot to feel less-life-like.  Participant counts illustrated the prevalence of this experience, with nine participants noting disconnection at the baseline and six describing overheating at the highest setting. Across all conditions, participants emphasized the importance of realistic warmth, even heat distribution, and contextual suitability (e.g., cold rooms vs. warm environments). These themes provide valuable design guidance: moderate, biologically plausible warmth enhances perceived life-likeness and affective interaction, while excess or inconsistency can disrupt the illusion of a living presence.

The qualitative analysis further suggests that perceived realism is not solely a matter of temperature intensity, but also of coherence:  users expect warmth to be distributed in a way that is internally consistent and anatomically plausible. This feedback points to a key limitation of the current prototype: the localized concentration of heat likely diminished the plausibility of the thermal augmentation and, in turn, reduced the sense of believability. Future designs may benefit from more evenly distributed or multi-site thermal feedback to  convincingly replicate the full-body warmth of a living creature.

\section{Vibrotactile Interface}

For the design of the vibrotactile interface, we used three HapCoil-One\footnote{\url{https://tactilelabs.com/wp-content/uploads/2023/11/HapCoil_One_datasheet.pdf}} actuators controlled by a HSD-mk2 board developed by Actronika. The actuators were embedded inside the robot to resemble more closely the feeling of sensing an animal's purring or heartbeat. Two of them were located on each flank of the PARO robot, positioned between its actual fur and the thermal layer, to reproduce the heartbeat cue, creating a sensation of the heartbeat coming from within the robot's body rather than being surface-mounted or externally attached. These were positioned specifically in areas where a user might naturally expect to feel a heartbeat when touching a living being. An additional one was instead placed internally, in the throat area, to reproduce the purring cue.  This location was chosen to mimic the typical anatomical region where purring vibrations are felt in animals such as cats, while also ensuring the actuator remained securely in place despite the robot’s limb movements. Furthermore, we wrapped the actuator in a thin felt layer to soften any potential contact with hard internal surfaces and avoid distortion of the vibratory sensation. The internal placement created the impression that the vibration was emanating organically from within the robot itself.  The actuators are shown in Fig.~\ref{fig:thermal_interface}.

To ensure a coherent multisensory experience, we calibrated the intensity of each actuator so that the vibrations remained perceptible yet localized. Users can feel the vibratory cues in the immediate area surrounding each vibration motor, without the sensation diffusing throughout the entire robot. This helped maintain the illusion of life-like internal rhythms and contributed to the subtlety and believability of the tactile experience. All  cables were discreetly routed and merged with the main power cable of the heating pad, preserving the visual integrity of the prototype and minimizing distractions during user interaction.

The complete prototype, including thermal hardware (not used in the vibration study) and vibration actuators can be seen in  Fig.~\ref{fig:multimodal_thermal_prototype}. This cohesively integrates both feedback modalities  to support an emotionally engaging, life-like interaction with PARO.

\subsection{Tactile Cue Design}

Building from the results and limitations in~\cite{borgstedt2024soothing}, the heartbeat cue was refined to be more biologically plausible, while a purring cue was added to increase the diversity of haptic cues, diverging from the uni-dimensional approach presented in \cite{borgstedt2024soothing}. Albeit not identical to cat purring, seals have been observed to use purring-like vocalisations \cite{PozoGalvan2024}, making such a cue both biologically plausible and potentially intuitive for users, who are likely familiar with similar tactile and auditory patterns from interactions with domestic cats. This familiarity may enhance cue recognisability and emotional resonance, thereby aiming to improve user engagement. These cues were developed using a live generation approach through an iterative prototyping phase and implemented in \texttt{Max/MSP 8}\footnote{\url{https://cycling74.com/products/max}}, a programming environment used for real-time audio synthesis~\cite{manzo2016_max_user_guide}. The code used for the cue synthesis in Max can be found in the supplementary materials. Note that while such an environment provides more freedom on the design of the audio wave forms, it is not a system requirement as any audio file can be natively converted into vibration by the HSD control board.

The cues are defined as follows:
\begin{itemize}
    \item \textbf{Heartbeat Cue:} The baseline cue represented a resting heartbeat at 55~bpm and was based on parameters outlined by~\cite{Shim2020};
    \item \textbf{Purring Cue:} The purring cue featured 70~\textit{purrs-per-minute} and was based on a spectral analysis of audio recordings of cat purring. Through iterative testing, a purring cue was selected which consisted of a combination of 35~Hz and 60~Hz sine waves. Then, an ADSR envelope was applied to the 60~Hz component (A: 200~ms, D: 100~ms, S: 0.9, R: 250~ms). The envelope was triggered every $\frac{60}{70}$~s, with the release stage triggered every $\frac{60}{2(70)}$~s. The 35~Hz component represented the underlying vibration in the purr, with the 60~Hz component representing the rhythmic component from breathing. The resulting purring wave form $p(t)$ is expressed by:
    \begin{equation}
        p(t) = \sin (2\pi \cdot 35t) + A(t) \sin (2\pi \cdot 60t).
    \end{equation}
\end{itemize}

The cues were subsequently refined by varying three different dimensions that would influence the signal. For our refinement process these dimensions were either tuned to LOW (0-baseline) or to HIGH (1-maximum). However, we designed the tuning parameter to be in a continuous range of $[0,1]$ so that the dimensions could be set more granularly in the future for greater customization. The three dimensions were:
\begin{itemize}
    \item \textbf{Amplitude:} Controls the amplitude level of vibrotactile stimuli between 0\% and 100\% of the actuators' capacity. When Final Amplitude was LOW, the heartbeat and purring actuators were set to 66\% of maximum output. When Final Amplitude was HIGH, the heartbeat actuators were set to 75\% of maximum output and the purring actuator to 70\% of maximum output. Two separate values were used as the purring was felt more strongly than the heartbeat, and so a greater increase in amplitude was required for a similar perceptual increase in strength;
    \item \textbf{Variability:} Controls how closely the cue mimics variations in biological rhythms, by adding variation to the beats-per-minute and purrs-per-minute for the heartbeat and purring, and by adding variation in signal amplitude. With Variability LOW, the heartbeat was presented at 55 beats-per-minute, with the signal attenuated to 75\%. The purring was presented at 70 purrs-per-minute, with amplitude attenuated to 80\%. With Variability HIGH, heartbeat beats-per-minute varied sinusoidally by $\pm$ 25\% from the 55 beats-per-minute baseline over 30 seconds, ranging between 42 and 68 beats-per-minute. Each heartbeat was presented with a randomly chosen amplitude between 75\% and 100\%. Over the same 30 second sinusoidal period, the purring amplitude varied between 80\% and 100\%, and the purrs-per-minute varied by $\pm$ 25\% (52.5 to 87.5 purrs-per-minute);
    \item \textbf{Hyper-Realism:} Controls the `exaggeration' of the cue by boosting bass frequencies at 45~Hz. Exaggeration like this is often used in sound design to create signals which can sound more real or believable by simulating human expectations, rather than objective reality \cite{chion2019audiovision}. With hyper-realism HIGH, the purring cue is boosted by a 10.5~dB gain at 45~Hz, while the heartbeat one is boosted by a 9~dB gain at 45~Hz. A larger boost was required for the purring cue to achieve the desired characteristic.
\end{itemize}

These cues were used as baseline for an iterative refinement process carried out with experts.

\begin{table*}[th]
\aboverulesep=0ex 
\belowrulesep=0ex 
\centering
\caption{Summary of expert evaluations for vibrotactile settings based on perceived life-likeness.}
\resizebox{\textwidth}{!}{%
\begin{tabular}{ccccc}
\toprule
\makecell{\textbf{Life-likeness} \\ \textbf{Rating}} & 
\makecell{\textbf{Setting Specification} \\ \textbf{(Amplitude, Variability,} \\ \textbf{Hyper-Realism)}} & 
\makecell{\textbf{General Perception}} & 
\makecell{\textbf{Key Expert Feedback}} & 
\makecell{\textbf{Setting}} \\
\midrule
High & (LOW, HIGH, LOW) & Most life-like & \makecell{Subtle, soft, and internal; perceived as natural heartbeat \\ vibration felt embedded in robot body} & 4 \\
\midrule
Medium–High & (LOW, HIGH, HIGH) & Mixed–positive & \makecell{Realistic pulse pattern but slightly too strong;\\ described as more suitable for a larger animal} & 5 \\
\midrule
Medium & (LOW, LOW, HIGH) & Mixed–positive & \makecell{Smooth but ``buzzy''; felt more like a general vibration\\ than a heartbeat; lacked organic quality} & 6 \\
\midrule
Low–Medium & (HIGH, LOW, LOW) & Weak but soft & \makecell{Perceived as subtle and human-like but \\ often too faint to detect during movement} & 8 \\
\midrule
Low & (HIGH, LOW, HIGH) & Least life-like & \makecell{Overly intense and mechanical; \\ described as cartoon-like, artificial, and attention-grabbing} & 3 \\
\midrule
Low & (HIGH, HIGH, HIGH) & Least life-like & \makecell{Coarse and jarring; felt exaggerated and ``griddy'', \\ similar to a sewing machine or cartoon heartbeat} & 7 \\
\midrule
Low & (HIGH, HIGH, LOW) & Low realism & \makecell{Crunchy and inconsistent; reminded participants of\\ robotic or electronic components} & 2 \\
\midrule
Baseline & (LOW, LOW, LOW) & No vibration (baseline) & \makecell{Comfortable, but lacked perceived internal activity;\\ did not evoke a sense of presence or life} & 1 \\
\bottomrule
\end{tabular}
}
\label{tab:vibration_lifelikeness}
\end{table*}

\subsection{Expert Design Sessions}

The objective of the expert design sessions was to create life-like or emotionally resonant vibrotactile cues that would overcome the ``mechanical buzzing'' critique outlined by participants in~\cite{borgstedt2024soothing}. To this end, 5 individual co-design sessions with affective haptics research experts (4 men, 1 woman)  from our university~(all PhD holders and active in the field) were conducted. Experts volunteered their time without any additional incentives.
During these sessions, experts placed the PARO robot in their lap, positioning their hands close to the vibrotactile actuators, depending on the cue under refinement. Both cues with different parameter combinations (e.g. baseline cue with variability vs baseline cue without variability) were evaluated to understand how specific parameter combinations influenced perceived life-likeness. Each session followed the following protocol:
\begin{enumerate}
    \item \textbf{Introduction:} Experts were briefed on the study goals and what ``life-like'' tactile cues meant in the context of this study was explained;
    \item \textbf{Structured Comparisons:} The experts evaluated both vibrotactile cues which varied systematically along the three dimensions: \textit{Amplitude} (66\% baseline amplitude, 75\% and 80\% maximum amplitude for purring and heartbeat respectively), \textit{Variability} (LOW/HIGH variability), and \textit{Hyper-Realism} (LOW/ HIGH hyper-realism). Each stimulus was presented for approximately 30 seconds and a pairwise comparison approach was used to identify which settings felt more life-like while remaining comfortable. Audio recordings of each stimulus can be found in the supplementary materials;
    \item \textbf{Refinement:} Based on their own rankings, the expert selected the most and least life-like settings. Experts explained why specific settings felt more natural and suggested improvements, enabling \textit{in situ} co-design of the vibrotactile stimuli. 
\end{enumerate}

In practical terms, this meant working with the base heart rate of 55 beats-per-minute and base purr rate of 70 purrs-per-minute that provided a sensation of PARO being at rest. The experts evaluated how different settings along the \textit{Variability} and \textit{Hyper-Realism} dimensions affected the life-likeness of the vibrotactile feedback, ultimately identifying optimal configurations for enhancing users' emotional connection with PARO. The expert design sessions yielded critical insights for enhancing the life-likeness of vibrotactile heartbeat and purring cues, which are reported  in Table~\ref{tab:vibration_lifelikeness}. 

\subsection{Implementation of Final Vibrotactile Cues}

\begin{figure}
    \centering
    \includegraphics[width=.9\linewidth]{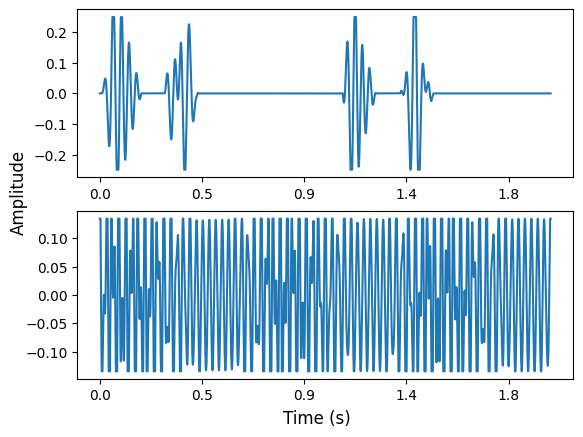}
    \caption{Waveforms of the final vibrotactile cues. Top: heartbeat. Bottom: purring.}
    \label{fig:realismwaveforms}
\end{figure}

Drawing on expert feedback, we implemented refined heartbeat and purring cues that took into account perceptibility, biological plausibility, and user comfort. Both cues were based on \textit{Setting 4}, which experts unanimously identified as the most life-like. This configuration retained the base frequency characteristics of the original cues but introduced natural Variability to enhance realism without overwhelming the user. The following paragraphs detail the final implementation of each cue, with plots shown in Figure~\ref{fig:realismwaveforms}.

\subsubsection*{\textbf{Heartbeat Cue}}
The final heartbeat cue simulated cardiac physiology through four precisely timed components:
\begin{itemize}
    \item Two atrial components: 25~Hz and 30~Hz sine pulses with duration  proportional to the beat interval ($\frac{2n}{21.33}$~ms and $\frac{2n}{12.8}$~ms respectively, with $n$ the interval between beats);
    \item Two ventricular components: 20~Hz and 30~Hz sine pulses (durations $\frac{2n}{12.8}$~ms and $\frac{2n}{14.2}$~ms), both triggered $\frac{n}{4}$~ms after the initial heartbeat onset;
    \item Variability: Random amplitude variations (75-100\%) and sinusoidal heart rate modulation (41.25-68.75~bpm over 30 seconds) were applied to mimic natural physiological fluctuations.
\end{itemize}

The final heartbeat cue $hb(t)$ is expressed by:
\begin{equation}
     hb(t) = A_r(Atr_1 + Atr_2 + Ven_1 + Ven_1),
\end{equation}
where:
\begin{equation*}
     Atr_1 = \sin(2\pi \cdot 25t)\ \text{for } 0 \leq t < \frac{120}{21.33 \cdot \text{bpm}},
\end{equation*}
\begin{equation*}
     Atr_2 = \sin(2\pi \cdot 30t)\ \text{for } 0 \leq t < \frac{120}{12.8 \cdot \text{bpm}},
\end{equation*}
\begin{equation*}
     Ven_1 = \sin(2\pi \cdot 20t)\ \text{for } \frac{60}{4 \cdot \text{bpm}} \leq t < \frac{120}{12.8 \cdot \text{bpm}},
\end{equation*}
\begin{equation*}
     Ven_2 = \sin(2\pi \cdot 30t)\ \text{for } \frac{60}{4 \cdot \text{bpm}} \leq t < \frac{120}{14.2 \cdot \text{bpm}},
\end{equation*}
and:

\begin{equation*}
    A_r \sim U(0.75,1), \quad bpm = bpm_{base} \left[ 1 + 0.25 \sin \left( \frac{2\pi}{30}t \right) \right].
\end{equation*}

\subsubsection*{\textbf{Purring Cue}}
The final purring cue combined two frequency components to create a biomimetic simulation:
\begin{itemize}
\item \textit{Base vibration:} A continuous 35~Hz sine wave providing a steady, low-frequency foundation.
\item \textit{Purr articulation:} An envelope-modulated 60~Hz sine wave generating individual purr sensations within a 550~ms cycle (200~ms ramp-up, 100~ms subtle decrease, 250~ms ramp-down).
\item \textit{Variability:} Amplitude modulation (80–100\%) and dynamic purr rate fluctuation (52.5–87.5 purrs-per-minute) were incorporated to reflect the natural variability of feline purring.
\end{itemize}

The final purring cue $p(t)$ is expressed by:
\begin{equation}
    p(t) = V(t) \left[ \sin(2\pi \cdot 35t) + A(t)\sin(2\pi \cdot 60t) \right],
\end{equation}

where:

\begin{equation*}
    V(t) = 0.9 + 0.1 \sin \left( \frac{2\pi}{30}t \right)
\end{equation*}

 and
\begin{equation*}
    A(t) = \begin{cases}
\frac{t}{0.2} & \text{for } 0 \leq t < 0.2 \\
1 - (t - 0.2) & \text{for } 0.2 \leq t < 0.3 \\
0.9 & \text{for } 0.3 \leq t < k(t)\\
0.9 - \frac{0.9}{0.25}(t - k(t)) & \text{for } k(t) \leq t < k(t) + 0.25 \\
0 & \text{for } t \geq k(t) + 0.25
\end{cases},
\end{equation*}

with

\begin{equation*}
    k(t) = \frac{60}{2 \cdot ppm}, \quad ppm = 70 \left[ 1 + 0.25 \sin \left( \frac{2\pi}{30}t \right) \right].
\end{equation*}

\section{Conclusions}

This paper outlined the design process of enhancing a SAR with bio-inspired haptic cues. The final prototype (Figure~\ref{fig:multimodal_thermal_prototype}) is an augmented version of an 8th-generation \textit{PARO} that combines a commercially available heating pad and embedded tactile actuators to deliver biologically plausible lifelike haptic feedback. Thermal stimulation is provided by a flexible electric pad encased in synthetic white fur that matches PARO’s original surface and wraps around its torso, maintaining tactile fidelity and zoomorphic appearance. A temperature of 35-36\,\textdegree C offers an optimal balance between comfort and biological plausibility. Vibrotactile feedback is generated by three strategically placed actuators: one in the “throat” region producing a continuous purring effect, and two along the flanks simulating heartbeat-like rhythms. These locations correspond to anatomical sites where such sensations are typically perceived, enhancing biological plausibility. All cabling is routed internally and integrated with the thermal system’s power supply, resulting in a seamless, unobtrusive design.  In summary, this work makes four distinct contributions to the field of human-robot interaction:

\begin{enumerate}
    \item \textbf{Extending an evidence-based platform into a versatile testbed.} Rather than creating a new device we augment \textit{PARO}, a SAR with a proven wellbeing record, and show how such an approach expands its utility. Because both thermal and vibrotactile parameters can be readily adjusted, the system serves as a flexible platform for future studies on user preferences, group-specific adaptations, and deployment across care, home, or educational settings.

    \item \textbf{A reproducible vibrotactile design pipeline.} We specify waveform structures and adjustable design parameters (amplitude, variability, hyper-realism) for heartbeat- and purring-like cues. This methodological transparency enables replication, supports cross-platform adaptation, and facilitates iterative refinement of tactile feedback.

    \item \textbf{User-informed thermal design guidelines.} Through collecting data on user preferences and perceptions, we identify temperature ranges users perceive as both believable and comfortable ($\simeq$35-36\,\textdegree C), and demonstrate how warmth intensity and spatial distribution shape perceptions of realism. These findings provide empirically grounded guidance for future thermal implementations.

    \item \textbf{Low-cost, practical integration guidance.} We document component selection, anatomical placement, cabling strategies, and hygiene considerations (including a washable thermal cover). 
\end{enumerate}

\noindent Together, these contributions demonstrate how careful, user-centred multimodal design can transform a widely deployed SAR into a versatile experimental platform: one that enriches embodied interaction and supports sustained exploration of affective touch, user adaptation, and deployment in care settings.

\bibliographystyle{IEEEtran}
\bibliography{references}

@article{PozoGalvan2024,
  author    = {Pozo Galván, Y. P. and Pérez Tadeo, M. and Pommier, M. and O'Brien, J.},
  title     = {Static Acoustic Monitoring of Harbour (*Phoca vitulina*) and Grey Seals (*Halichoerus grypus*) in the Malin Sea: A Revolutionary Approach in Pinniped Conservation},
  journal   = {Journal of Marine Science and Engineering},
  year      = {2024},
  volume    = {12},
  number    = {1},
  pages     = {118},
  doi       = {10.3390/jmse12010118},
  url       = {https://doi.org/10.3390/jmse12010118}
}

@book{chion2019audiovision,
  author    = {Chion, Michel and Gorbman, Claudia},
  title     = {Audio-Vision: Sound on Screen},
  year      = {2019},
  publisher = {Columbia University Press},
  address   = {New York}
}

@INPROCEEDINGS{borgstedt2022hot,
 author={Borgstedt, J. and Pollick, F.E. and Brewster, S.},
 booktitle={2022 31st IEEE International Conference on Robot and Human Interactive Communication (RO-MAN)}, 
 title={Hot or not? Exploring User Perceptions of thermal Human-Robot Interaction}, 
 year={2022},
 volume={},
 number={},
 pages={1334-1340},
 doi={10.1109/RO-MAN53752.2022.9900785}
}

@INPROCEEDINGS{borgstedt2024soothing,
   author={Borgstedt, Jacqueline and Macdonald, Shaun and Marky, Karola and Pollick, Frank E. and Brewster, Stephen A.},
 booktitle={2024 33rd IEEE International Conference on Robot and Human Interactive Communication (ROMAN)}, 
  title={Soothing Sensations: Enhancing Interactions with a Socially Assistive Robot through Vibrotactile Heartbeats}, 
   year={2024},
  volume={},
  number={},
  pages={231-237},
  doi={10.1109/RO-MAN60168.2024.10731372}
}

@book{manzo2016_max_user_guide,
  title={Max/MSP/Jitter for music: A practical guide to developing interactive music systems for education and more},
  author={Manzo, Vincent J},
  year={2016},
  publisher={Oxford University Press}
}

@inproceedings{Shim2020,
   author = {Sang Won Shim and Hong Z. Tan},
   doi = {10.1007/978-3-030-49713-2_37},
   isbn = {9783030497125},
   issn = {16113349},
   journal = {Lecture Notes in Computer Science (including subseries Lecture Notes in Artificial Intelligence and Lecture Notes in Bioinformatics)},
   pages = {532-548},
   publisher = {Springer},
   title = {palmscape: Calm and pleasant vibrotactile signals},
   volume = {12200 LNCS},
   year = {2020},
}

@inproceedings{Elkin2021,
   author = {Lisa A. Elkin and Matthew Kay and James J. Higgins and Jacob O. Wobbrock},
   doi = {10.1145/3472749.3474784},
   journal = {UIST 2021 - Proceedings of the 34th Annual ACM Symposium on User Interface Software and Technology},
   title = {An Aligned Rank Transform Procedure for Multifactor Contrast Tests},
   year = {2021},
}

@inproceedings{Wobbrock2011,
   author = {Jacob O. Wobbrock and Leah Findlater and Darren Gergle and James J. Higgins},
   doi = {10.1145/1978942.1978963},
   journal = {Conference on Human Factors in Computing Systems - Proceedings},
   title = {The Aligned Rank Transform for nonparametric factorial analyses using only ANOVA procedures},
   year = {2011},
}

@article{Braun2006,
   author = {V Braun and V Clarke - Qualitative research in psychology and undefined 2006},
   doi = {10.1191/1478088706qp063oa},
   issue = {2},
   journal = {Taylor  \&  Francis},
   pages = {77-101},
   title = {Using thematic analysis in psychology},
   volume = {3},
   year = {2006},
}

@article{OConnor2021,
   author = {Daryl B. O'Connor and Julian F. Thayer and Kavita Vedhara},
   issn = {15452085},
   journal = {Annual review of psychology},
   month = {1},
   pages = {663-688},
   pmid = {32886587},
   publisher = { Annual Reviews },
   title = {Stress and Health: A Review of Psychobiological Processes},
   volume = {72},
   year = {2021},
}

@article{Shields2017,
   author = {Grant S. Shields and George M. Slavich},
   doi = {10.1111/SPC3.12335},
   issn = {1751-9004},
   issue = {8},
   journal = {Social and Personality Psychology Compass},
   month = {8},
   pages = {e12335},
   publisher = {John Wiley \& Sons, Ltd},
   title = {Lifetime stress exposure and health: A review of contemporary assessment methods and biological mechanisms},
   volume = {11},
   year = {2017},
}

@article{Dagnino-Subiabre2022,
   author = {Alexies Dagnino-Subiabre},
   doi = {10.1016/J.COBEHA.2021.08.011},
   issn = {2352-1546},
   journal = {Current Opinion in Behavioral Sciences},
   month = {2},
   pages = {75-79},
   publisher = {Elsevier},
   title = {Resilience to stress and social touch},
   volume = {43},
   year = {2022},
}

@article{Jakubiak2019,
   author = {Brittany K. Jakubiak and Brooke C. Feeney},
   doi = {10.1177/0146167218788556},
   issn = {15527433},
   issue = {3},
   journal = {Personality and Social Psychology Bulletin},
   title = {Hand-in-Hand Combat: Affectionate Touch Promotes Relational Well-Being and Buffers Stress During Conflict},
   volume = {45},
   year = {2019},
}

@article{Coan2006,
   author = {James A. Coan and Hillary S. Schaefer and Richard J. Davidson},
   doi = {10.1111/j.1467-9280.2006.01832.x},
   issn = {09567976},
   issue = {12},
   journal = {Psychological Science},
   title = {Lending a hand: Social regulation of the neural response to threat},
   volume = {17},
   year = {2006},
}

@article{Nimer2007,
author = {Nimer, Janelle and Lundahl, Brad},
doi = {10.2752/089279307X224773},
eprint = {NIHMS150003},
isbn = {0892-7936},
issn = {08927936},
journal = {Anthrozoos},
number = {3},
pages = {225--238},
pmid = {21729249},
title = {{Animal-assisted therapy: A meta-analysis}},
volume = {20},
year = {2007}
}

@article{Eckstein2020_touch_stress,
   author = {Monika Eckstein and Ilshat Mamaev and Beate Ditzen and Uta Sailer},
   doi = {10.3389/FPSYT.2020.555058/FULL},
   issn = {16640640},
   journal = {Frontiers in Psychiatry},
   month = {11},
   pages = {555058},
   publisher = {Frontiers Media S.A.},
   title = {Calming Effects of Touch in Human, Animal, and Robotic Interaction—Scientific State-of-the-Art and Technical Advances},
   volume = {11},
   year = {2020},
}

@article{Haggerty2017,
   author = {Julie M. Haggerty and Megan Kiely Mueller},
   doi = {10.1007/s10755-017-9392-0},
   issn = {15731758},
   issue = {5-6},
   journal = {Innovative Higher Education},
   title = {Animal-assisted Stress Reduction Programs in Higher Education},
   volume = {42},
   year = {2017},
}

@article{Chen2019,
   author = {Shu-Chuan Chen and Wendy Moyle and Cindy Jones},
   doi = {10.1093/GERONI/IGZ038.1236},
   issue = {Suppl 1},
   journal = {Innovation in Aging},
   month = {11},
   pages = {S341},
   publisher = {Oxford University Press},
   title = {THE IMPACT OF A PARO INTERVENTION ON DEPRESSION AND WELL-BEING IN OLDER ADULTS WITH DEPRESSION},
   volume = {3},
   year = {2019},
}

@article{packheiser2024_touch_interventions,
  title={A systematic review and multivariate meta-analysis of the physical and mental health benefits of touch interventions},
  author={Packheiser, Julian and Hartmann, Helena and Fredriksen, Kelly and Gazzola, Valeria and Keysers, Christian and Michon, Fr{\'e}d{\'e}ric},
  journal={Nature human behaviour},
  volume={8},
  number={6},
  pages={1088-1107},
  year={2024},
  publisher={Nature Publishing Group UK London}
}

@INPROCEEDINGS{seifer2005_social_robotics_definition,
  author={Feil-Seifer, D. and Matarić, M.J.},
  booktitle={9th International Conference on Rehabilitation Robotics, 2005. ICORR 2005.}, 
  title={Defining socially assistive robotics}, 
  year={2005},
  pages={465-468},
  doi={10.1109/ICORR.2005.1501143}
}

@article{Geva2020,
   author = {Nirit Geva and Florina Uzefovsky and Shelly Levy-Tzedek},
   doi = {10.1038/s41598-020-66982-y},
   issn = {20452322},
   issue = {1},
   journal = {Scientific Reports},
   month = {12},
   pages = {1-15},
   pmid = {32555432},
   publisher = {Nature Research},
   title = {Touching the social robot PARO reduces pain perception and salivary oxytocin levels},
   volume = {10},
   year = {2020},
}

@article{Wada2005,
   author = {Kazuyoshi Wada and Takanori Shibata and Tomoko Saito and Kayoko Sakamoto and Kazuo Tanie},
   doi = {10.1109/ROBOT.2005.1570535},
   isbn = {078038914X},
   issn = {10504729},
   journal = {Proceedings - IEEE International Conference on Robotics and Automation},
   pages = {2785-2790},
   title = {Psychological and social effects of one year robot assisted activity on elderly people at a health service facility for the aged},
   volume = {2005},
   year = {2005},
}

@article{Aminuddin2017,
   author = {Raihah Aminuddin and Amanda Sharkey},
   doi = {10.1145/3121283.3121420},
   isbn = {9781450352567},
   journal = {Proceedings of the European Conference on Cognitive Ergonomics},
   month = {9},
   pages = {63-64},
   publisher = {Association for Computing Machinery},
   title = {A Paro robot reduces the stressful effects of environmental noise},
   year = {2017},
}

@article{Jeong2015,
   author = {Sooyeon Jeong and Deirdre E. Logan and Matthew S. Goodwin and Suzanne Graca and Brianna O'Connell and Honey Goodenough and Laurel Anderson and Nicole Stenquist and Katie Fitzpatrick and Miriam Zisook and Luke Plummer and Cynthia Breazeal and Peter Weinstock},
   doi = {10.1145/2701973.2702028},
   isbn = {9781450333184},
   issn = {21672148},
   journal = {ACM/IEEE International Conference on Human-Robot Interaction},
   month = {3},
   pages = {103-104},
   publisher = {IEEE Computer Society},
   title = {A Social Robot to Mitigate Stress, Anxiety, and Pain in Hospital Pediatric Care},
   volume = {02-05-March-2015},
   year = {2015},
}

@article{Hung2019_paro_review,
   author = {Lillian Hung and Cindy Liu and Evan Woldum and Andy Au-Yeung and Annette Berndt and Christine Wallsworth and Neil Horne and Mario Gregorio and Jim Mann and Habib Chaudhury},
  title     = {The Benefits of and Barriers to Using a Social Robot PARO in Care Settings: A Scoping Review},
  journal   = {BMC Geriatrics},
  volume    = {19},
  number    = {1},
  pages     = {232},
  year      = {2019},
  doi       = {10.1186/s12877-019-1244-6},
  issn      = {1471-2318}
}

@article{Bonarini2016,
   author = {Andrea Bonarini and Franca Garzotto and Mirko Gelsomini and Maximiliano Romero and Francesco Clasadonte and Ayse Naciye Celebi Yilmaz},
   doi = {10.1109/ROMAN.2016.7745214},
   isbn = {9781509039296},
   journal = {25th IEEE International Symposium on Robot and Human Interactive Communication, RO-MAN 2016},
   month = {11},
   pages = {823-830},
   publisher = {Institute of Electrical and Electronics Engineers Inc.},
   title = {A huggable, mobile robot for developmental disorder interventions in a multi-modal interaction space},
   year = {2016},
}

@article{Saldien2008,
   author = {Jelle Saldien and Kristof Goris and Selma Yilmazyildiz and Werner Verhelst and Dirk Lefeber},
   doi = {10.14198/JOPHA.2008.2.2.02},
   issn = {18880258},
   issue = {2},
   journal = {Journal of Physical Agents},
   pages = {3-11},
   publisher = {University of Alicante},
   title = {On the Design of the Huggable Robot Probo},
   volume = {2},
   year = {2008},
}

@article{Bates2020,
   author = {Danika Passler Bates and James E Young},
   city = {New York, NY, USA},
   doi = {10.1145/3406499},
   isbn = {9781450380546},
   journal = {Proceedings of the 8th International Conference on Human-Agent Interaction},
   pages = {pages},
   publisher = {ACM},
   title = {SnuggleBot: A Novel Cuddly Companion Robot Design},
   volume = {3},
   year = {2020},
}

@inproceedings{Block2021,
   author = {Alexis E. Block and Sammy Christen and Roger Gassert and Otmar Hilliges and Katherine J. Kuchenbecker},
   doi = {10.1145/3434073.3444656},
   issn = {21672148},
   journal = {ACM/IEEE International Conference on Human-Robot Interaction},
   title = {The six hug commandments: Design and evaluation of a human-sized hugging robot with visual and haptic perception},
   year = {2021},
}

@article{hayward2004_haptics,
  title={Haptic interfaces and devices},
  author={Hayward, Vincent and Astley, Oliver R and Cruz-Hernandez, Manuel and Grant, Danny and Robles-De-La-Torre, Gabriel},
  journal={Sensor review},
  volume={24},
  number={1},
  pages={16--29},
  year={2004},
  publisher={Emerald Group Publishing Limited}
}

@article{rantala2013touchgestures,
  author    = {Rantala, Jussi and Salminen, Katri and Raisamo, Roope and Surakka, Veikko},
  title     = {Touch Gestures in Communicating Emotional Intention via Vibrotactile Stimulation},
  journal   = {International Journal of Human-Computer Studies},
  volume    = {71},
  number    = {6},
  pages     = {679--690},
  year      = {2013},
  issn      = {1071-5819},
  doi       = {10.1016/j.ijhcs.2013.02.004},
}

@inproceedings{Ju2021emotioncommunicationvibro,
author = {Ju, Yulan and Zheng, Dingding and Hynds, Danny and Chernyshov, George and Kunze, Kai and Minamizawa, Kouta},
title = {Haptic Empathy: Conveying Emotional Meaning through Vibrotactile Feedback},
year = {2021},
isbn = {9781450380959},
publisher = {Association for Computing Machinery},
address = {New York, NY, USA},
doi = {10.1145/3411763.3451640},
booktitle = {Extended Abstracts of the 2021 CHI Conference on Human Factors in Computing Systems},
articleno = {225},
numpages = {7},
location = {Yokohama, Japan},
series = {CHI EA '21}
}

@inproceedings{Wilson2017,
   author = {Graham Wilson and Stephen A. Brewster},
   city = {New York, NY, USA},
   doi = {10.1145/3025453.3025614},
   isbn = {9781450346559},
   journal = {Conference on Human Factors in Computing Systems - Proceedings},
   month = {5},
   pages = {1743-1755},
   publisher = {Association for Computing Machinery},
   title = {Multi-Moji: Combining thermal, vibrotactile  \&  visual stimuli to expand the affective range of feedback},
   volume = {2017-May},
   url = {https://dl.acm.org/doi/10.1145/3025453.3025614},
   year= {2017},
}

@article{dobrushina2024interoceptive,
  title={Interoceptive training with real-time haptic versus visual heartbeat feedback},
  author={Dobrushina, Olga and Tamim, Yossi and Wald, Iddo Yehoshua and Maimon, Amber and Amedi, Amir},
  journal={Psychophysiology},
  volume={61},
  number={11},
  pages={e14648},
  year={2024},
  publisher={Wiley Online Library}
}

@INPROCEEDINGS{zhou2020_heartbeat_haptics,
  author={Zhou, Yizhen and Murata, Aiko and Watanabe, Junji},
  booktitle={2020 IEEE Haptics Symposium (HAPTICS)}, 
  title={The Calming Effect of Heartbeat Vibration}, 
  year={2020},
  volume={},
  number={},
  pages={677-683},
  doi={10.1109/HAPTICS45997.2020.ras.HAP20.157.5a2e1551}
}

@article{bernardi2009entrainment,
  author    = {Bernardi, Luciano and Porta, Alberto and Casucci, Giuseppe and Balsamo, Rosita and Bernardi, Nicola Francesco and Fogari, Roberto and Sleight, Peter},
  title     = {Dynamic interactions between musical, cardiovascular, and cerebral rhythms in humans},
  journal   = {Circulation},
  volume    = {119},
  number    = {25},
  pages     = {3171--3180},
  year      = {2009},
  month     = {Jun},
  doi       = {10.1161/circulationaha.108.806174},
  pmid      = {19569263}
}

@article{hertenstein2002touch_heat,
  title={Touch: Its communicative functions in infancy},
  author={Hertenstein, Matthew J},
  journal={Human Development},
  volume={45},
  number={2},
  pages={70--94},
  year={2002},
  publisher={S. Karger AG Basel, Switzerland}
}

@article{ackerley2014_touch_warmth,
  title={Human C-tactile afferents are tuned to the temperature of a skin-stroking caress},
  author={Ackerley, Rochelle and Wasling, Helena Backlund and Liljencrantz, Jaquette and Olausson, H{\aa}kan and Johnson, Richard D and Wessberg, Johan},
  journal={Journal of Neuroscience},
  volume={34},
  number={8},
  pages={2879--2883},
  year={2014},
  publisher={Society for Neuroscience}
}

@article{Nie2012,
   author = {Jiaqi Nie and Michelle Pak and Angie Lorena Marin and S. Shyam Sundar},
   doi = {10.1145/2157689.2157755},
   isbn = {9781450310635},
   journal = {HRI'12 - Proceedings of the 7th Annual ACM/IEEE International Conference on Human-Robot Interaction},
   month = {4},
   pages = {201-202},
   title = {Can you hold my hand?: Physical warmth in human-robot interaction},
   year = {2012},
}

@article{Park2014,
   author = {Eunil Park and Jaeryoung Lee},
   doi = {10.1017/S026357471300074X},
   journal = {Robotica},
   pages = {133-142},
   publisher = {Cambridge University Press},
   title = {I am a warm robot: the effects of temperature in physical human-robot interaction},
   volume = {32},
   year = {2014},
}

@inproceedings{Nakanishi_warm_robot_hand,
    author = {Nakanishi, Hideyuki and Tanaka, Kazuaki and Wada, Yuya},
    title = {Remote handshaking: touch enhances video-mediated social telepresence},
    year = {2014},
    isbn = {9781450324731},
    publisher = {Association for Computing Machinery},
    address = {New York, NY, USA},
    doi = {10.1145/2556288.2557169},
    booktitle = {Proceedings of the SIGCHI Conference on Human Factors in Computing Systems},
    pages = {2143–2152},
    numpages = {10},
    location = {Toronto, Ontario, Canada},
    series = {CHI '14}
}

@article{Azevedo2017,
   author = {Ruben T. Azevedo and Nell Bennett and Andreas Bilicki and Jack Hooper and Fotini Markopoulou and Manos Tsakiris},
   doi = {10.1038/s41598-017-02274-2},
   issn = {2045-2322},
   issue = {1},
   journal = {Scientific Reports 2017 7:1},
   month = {5},
   pages = {1-7},
   pmid = {28536417},
   publisher = {Nature Publishing Group},
   title = {The calming effect of a new wearable device during the anticipation of public speech},
   volume = {7},
   year = {2017},
}

@inproceedings{Yohanan2011design,
   author = {Steve Yohanan and Karon E. MacLean},
   doi = {10.1145/1957656.1957820},
   journal = {HRI 2011 - Proceedings of the 6th ACM/IEEE International Conference on Human-Robot Interaction},
   title = {Design and assessment of the haptic creature's affect display},
   year = {2011},
}

\end{document}